%% file: root.tex
\documentclass[letterpaper, 10 pt, conference]{ieeeconf}  
\IEEEoverridecommandlockouts                              
\input{macros}

\title{\LARGE \bf 
Scalable Multi-modal Model Predictive Control via Duality-based Interaction Predictions

\vspace{-2mm}
}
\author{Hansung Kim*, Siddharth H. Nair*, Francesco Borrelli
\thanks{
HK, SHN, FB are with the Model Predictive Control Laboratory, UC Berkeley. E-mails: 
\{hansung, siddharth\_nair, fborrelli\}@berkeley.edu}
\vspace{-2mm}
}

\begin{document}

\maketitle
\thispagestyle{empty}
\pagestyle{empty}

\begin{abstract}
We propose a hierarchical architecture designed for scalable real-time Model Predictive Control (MPC) in complex, multi-modal traffic scenarios. This architecture comprises two key components: 1) RAID-Net, a novel attention-based Recurrent Neural Network that predicts relevant interactions along the MPC prediction horizon between the autonomous vehicle and the surrounding vehicles using Lagrangian duality, and 2) a reduced Stochastic MPC problem that eliminates irrelevant collision avoidance constraints, enhancing computational efficiency. Our approach is demonstrated in a simulated traffic intersection with interactive surrounding vehicles, showcasing a 12x speed-up in solving the motion planning problem. A video demonstrating the proposed architecture in multiple complex traffic scenarios can be found here: {\color{blue} \href{https://youtu.be/-pRiOnPb9_c}{link}}. GitHub: {\color{blue}\href{https://github.com/MPC-Berkeley/hmpc_raidnet}{link}}
\end{abstract}

\input{Sections/intro}

\input{Sections/problem_formulation}

\input{Sections/HMPC}

\input{Sections/traffic_intersection}
\input{Sections/results}

\input{Sections/conclusion}

\vspace{-1.0mm}
\bibliographystyle{IEEEtran}
\bibliography{references.bib}
\input{Sections/appendix}

\end{document}

%% file: macros.tex
\usepackage{verbatim}
\usepackage{graphics} 
\usepackage{amsmath,amssymb,mathtools,amsfonts}
\usepackage{multirow}
\usepackage{hhline}
\usepackage{algorithm}
\usepackage{footmisc}
\usepackage{flushend}
\usepackage[noend]{algpseudocode}
\usepackage{subcaption}
\usepackage{algorithmicx}
\usepackage{multirow}
\usepackage[utf8]{inputenc}
\usepackage{import}
\usepackage{graphicx, graphics}
\usepackage{caption}
\usepackage{subcaption}
\usepackage{array}
\usepackage{color}
\usepackage[table]{xcolor}
\usepackage{siunitx}
\usepackage[short]{optidef}
\usepackage{float}
\usepackage{cite}
\PassOptionsToPackage{hyphens}{url}\usepackage{hyperref}
\usepackage{bbold}
\definecolor{myedit}{rgb}{1.0,0,0}
\definecolor{mytodo}{rgb}{0,0.0,1.0}
\definecolor{lightgray}{rgb}{0.8, 0.8, 0.8}



%% file: Sections/intro.tex
\section{Introduction}
\label{sec:introduction}
Autonomous driving research on motion planning has graduated to address challenges in complex, interaction-driven urban scenarios over the past few years. In these settings, autonomous vehicles must navigate safely through highly uncertain environments, while observing and reacting to heterogeneous traffic agents: human-driven and autonomous vehicles navigating and making their own decisions. Accounting for these agents during motion planning presents intricate challenges, demanding robust and real-time solutions for safe navigation in urban environments. \par
 
aIn urban driving scenarios, the motion planning for autonomous vehicles in the presence of uncertain, multi-modal human-driven and autonomous vehicles poses a significant challenge, leading to the development of various solutions for planning and behavior prediction. These solutions can be categorized into three broad approaches: (i) Hierarchical Prediction and Planning \cite{salzmann2021trajectron, branchMPC, nair2023predictive}: where a sophisticated prediction architecture provides forecasts of the surrounding vehicles which is used for motion planning, (ii) Model-based Integrated Planning and Prediction \cite{espinoza2022deep,zhu2023sequential, peters2024contingency}: where planning and behavior prediction are simultaneously obtained by game-theoretic and joint-optimization approaches for all vehicles,  and (iii) End-to-End Learning-based Prediction and Control \cite{leurent2019social,hu2023planning}: where a sophisticated neural network (NN), trained using imitation/reinforcement learning algorithms on realistic datasets, implicitly and jointly forecasts the behavior of surrounding vehicles and a motion plan for the autonomous vehicle. Each of these approaches suffers from either scalability for complex driving scenarios or interpretability and safety of the computed motion plans. For instance in (i), \cite{nair2023predictive} employs Gaussian mixture models to explicitly express multi-modal predictions and positional uncertainty of surrounding vehicles for multi-modal motion planning. While this approach showcases robust navigation capabilities in multi-modal traffic scenarios, it focuses on interactions in simple traffic situations with limited traffic vehicles and does not scale for real-time applicability in complex scenarios with many surrounding vehicles and their multi-modal predictions. The game theoretic approaches in (ii) are generally computationally intractable for traffic scenarios with many vehicles/agents, which is further exacerbated when the games are multi-modal/mixed. The End-to-End approaches in (iii) are generally scalable to complex traffic scenarios but lack interpretability in their predictions. Given the traffic scene, it is unclear how to interpret which vehicles and which of their possible maneuvers are relevant for a safe motion plan for the autonomous vehicle (AV). In \cite{leurent2019social}, an attention-based architecture discerns relevant vehicles at the current time-step when controlling an autonomous vehicle in a simulated traffic intersection, but does not explicitly consider safety or long-term interactions. 
Hence, there is a pressing need for a scalable motion planning method that incorporates multi-modal predictions of surrounding vehicles while safely navigating complex interaction-driven scenarios for autonomous driving. \par


\begin{figure}
    \centering
    \includegraphics[width=0.9\columnwidth]{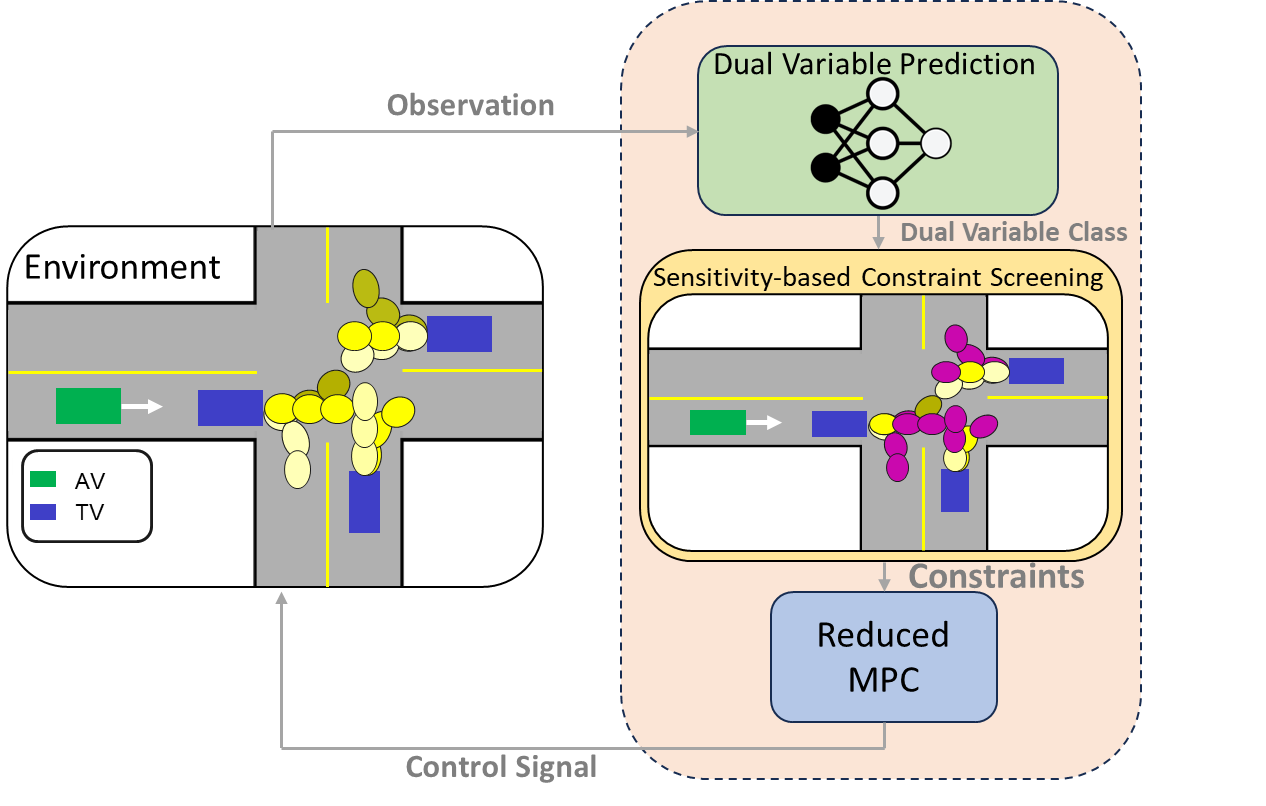}
    \caption{{\small{Our hierarchical architecture for motion planning with duality-based interaction prediction. Given the environment observation, we classify which \textit{vehicles and their maneuvers can be eliminated/screened} for solving a reduced, real-time MPC problem.}}}
    \label{fig:alg_architecture}
    \vspace{-0.7cm}
\end{figure}
In this work, we present a scalable real-time hierarchical motion planning algorithm given multi-modal predictions in complex, interaction-driven traffic scenarios. In the design stage we utilize an optimization-based motion planner and its Lagrangian dual to deduce which surrounding vehicle and its multi-modal predictions/maneuvers along the prediction horizon are \textit{interacting} with the autonomous vehicle. We design a Recurrent Neural Network (RNN) \cite{mandic2001recurrent} with attention mechanisms to predict the interactions between the autonomous vehicle and its surrounding vehicles. The RNN is trained from data from the optimization-based motion planner using a novel constraint screening criterion which uses ideas from sensitivity analysis as depicted in Fig. \ref{fig:alg_architecture}. The novel attention-based RNN architecture for interaction prediction is interpretable using Lagrange duality and generalizable to different traffic scenarios with varying numbers of surrounding vehicles. We demonstrate our proposed algorithm in numerical simulations of traffic intersections with interactive surrounding vehicles and demonstrate the real-time applicability of our motion planning approach. 
\par
The paper is structured as follows. Section \ref{sec:prblm_f} formulates Stochastic Model Predictive Control (MPC) for motion planning with multi-modal predictions, discussing scalability issues. In Section \ref{sec:methods}, we present a duality-based interpretation of the interaction between the autonomous vehicle and surrounding vehicles and introduce an RNN architecture for predicting interactions with multi-modal forecasts. A hierarchical algorithm for motion planning is outlined, utilizing NN predictions and a dual sensitivity-based constraint screening mechanism. Section \ref{sec:example} details the custom traffic intersection simulator used for training the NN through imitation learning and evaluating the proposed hierarchical motion planning algorithm. Section \ref{sec:results} evaluates our hierarchical MPC algorithm at the simulated traffic intersection. Finally, Section \ref{sec:conclusion} concludes the paper and discusses future research directions.

%% file: Sections/problem_formulation.tex
\section*{Notation}
The index set $\{k_1,k_1+1,\dots, k_2\}$ is denoted by $\mathbb{I}_{k_1}^{k_2}$. For a proper cone $\mathbb{K}$ and $x,y\in\mathbb{K}$, the dual cone of $\mathbb{K}$ is given by the convex set $\mathbb{K}^*=\{y| y^\top x\geq 0, \forall x\in\mathbb{K}\}$. $||\cdot||_p$ denotes the $p-$norm. For any matrix $A\in\mathbb{R}^{n\times m}$, we denote $[A]_\mathbb{S}$ to be the rows of $A$ with indices in $\mathbb{S}\subset\mathbb{I}_1^n$. For any pair of vectors $x,y \in \mathbb{R}^n$, we denote the Hadamard product as $x\circ y = [[x]_1[y]_1,..,[x]_n[y]_n]\in\mathbb{R}^n$.

\section{Problem Formulation}\label{sec:prblm_f}
\subsection{Vehicle Prediction Models}
Consider a controlled agent described by a linear time-varying discrete-time model
\small\begin{align}\label{eq:ev_dyn}
    x_{t+1}=A_tx_t+B_tu_t+E_tw_t
\end{align}\normalsize
where $x_t\in\mathbb{R}^{n_x}, u_t\in\mathbb{R}^{n_u}, w_t\in\mathbb{R}^{n_x}$ are the state, input and process noise respectively and $A_t,B_t,E_t$ are the system matrices at time $t$. The process noise $w_t$ is assumed to be a zero mean Gaussian, $\mathcal{N}(0, \Sigma_{t})$.

Now suppose that there are $V$ target vehicles (TVs), each described by \textit{multi-modal}, affine time-varying discrete-time predictions over a horizon $N$ at time $t$,
\small\begin{align}\label{eq:tv_dyn}
o^i_{k+1|t,j}&=T_{k|t,j}^io_{k|t,j}^i+q^i_{k|t,j}+E^i_{k|t,j}n^i_{k|t,j},\nonumber\\&~~~~~~~~~~~~\forall k\in\mathbb{I}_t^{t+N}, i\in\mathbb{I}_1^V, j\in\mathbb{I}_1^M 
\end{align}\normalsize
 where $o^i_{k|t,j},n^i_{k|t,j}\in\mathbb{R}^{n_o}$ are the position and
 process noise predictions for target vehicle $i$ in mode $j$ and time $t$, at the $k$th time step and $T^i_{k|t,j},q^i_{k|t,j},E^i_{k|t,j}$ are the time, mode-dependent system matrices. The process noise is assumed to be distributed as $n^{i}_{k|t,j}\sim\mathcal{N}(0,\Sigma^i_{k|t,j})$.

 Notice that each target vehicle prediction has $M$ modes, which amounts to a total of $M^V$ mode configurations of the target vehicles. We call each mode configuration a \textit{scenario} and define the function $\bar{j}:\mathbb{I}_1^V\times\mathbb{I}_1^{M^V}\rightarrow\mathbb{I}_1^{M}$ such that $\bar{j}(i,m)$ returns the mode of target vehicle $i$ in scenario $m$. 
 



 \subsection{Constraints}
 The autonomous vehicle is subject to polytopic state-input constraints given by $\mathbb{XU}_t=\{(x,u)|~ F^x_{t}x+ F^u_{t}u\leq f_{t}\}$, for time $t$ and affine collision avoidance constraints given as
 $\mathbb{CA}^i_{k|t,j}=\{(x,o)|~ L^{x,i}_{k|t,j}x+L^{o,i}_{k|t,j}o\leq l^i_{k|t,j}~\}$ for each vehicle $i$ in mode $j$ at time step $k$, which are obtained by appropriate inner-approximations of the free space as half spaces \cite{brudigam2021stochastic, baldini2016spacecraft, nair2022collision}.
 \subsection{Stochastic MPC formulation}\label{sec:mpc_approach}
We compute a state-feedback control $u_t=\pi(x_t,o_t)$ for the controlled by solving the following finite-horizon stochastic optimal control problem towards tracking a given reference $\{x^{\text{ref}}_k, u^{\text{ref}}_k\}_{k=t}^{t+N}$,
{\small{
\vspace{-2mm}
\begin{subequations}\label{opt:MPC_skeleton}
\begin{align}
 \min_{\substack{\boldsymbol{\theta}_t}}&\; \sum\limits_{m=1}^{M^V}\mathbb{E}\!\left[\sum\limits_{k=t}^{t+N-1}\!||Q(x_{k+1|t,m}\!-\!x^{\text{ref}}_k)||_2^2\!+\!||R(u_{k|t,m}\!-\!u^{\text{ref}}_k)||_2^2 \right]\label{eq:cost}\\
 \text{s.t. }&\quad x_{k+1|t,m}=A_{k}x_{k|t,m}+B_{k} u_{k|t,m}+E_kw_{k|t},\label{eq:ev_dyn}\\
&\quad o^i_{k+1|t,j}=T^i_{k|t,j}o^i_{k|t,j}+q^i_{k|t,j}+E^i_{k|t,j}n^i_{k|t,j},\label{eq:tv_mm_pred}\\
&\quad \mathbb{P}((x_{k|t,m},u_{k|t,m})\not\in\mathbb{XU}_k)\leq \epsilon,\label{eq: xu_constr}\\
&\quad\mathbb{P}((x_{k|t,m},o^i_{k|t,\bar{j}(i,m)})\not\in\mathbb{CA}^i_{k|t,\bar{j}(i,m)})\leq \epsilon,\label{eq:ca_constr}\\
&\quad u_{k|t,m}=h_{k|t}+\sum_{i=1}^V K^i_{k|t,\bar{j}(i,m)}o^i_{k|t,\bar{j}(i,m)},\label{eq:pol_param}\\
&\quad u_{t|t,m}=u_{t|t,1},~x_{t|t,m}=x_t,\ o_{t|t,j}=o_t,\label{eq:non_antic}\\
&\quad~\forall k\in\mathbb{I}_t^{t+N-1}, j\in\mathbb{I}_1^{M}, m\in\mathbb{I}_1^{M^V}, i\in\mathbb{I}_1^{V} \nonumber
\end{align}
\end{subequations}}}
where the decision variables 
{\small{\begin{align}\label{eq:pol_params}
\boldsymbol{\theta}_t=\{(h_{k|t}, \{\{K^i_{k|t,j}\}_{i=1}^V\}_{j=1}^M)\}_{k=t}^{t+N}
\end{align}}}parameterize the feedback policies \eqref{eq:pol_param} of the controlled agent.  The second term in \eqref{eq:pol_param} encodes state feedback for the obstacles' states, to allow the AV to \textit{react} to different realizations of the obstacles' trajectories, thus greatly enhancing the feasibility of the MPC problem (especially in the presence of large uncertainties). The Stochastic MPC (SMPC) is given by the optimal solution of \eqref{opt:MPC_skeleton} as {\small{
\begin{align}\label{eq:MPC}u_t=\pi_{\mathrm{MPC}}(x_t,o_t)=u^\star_{t|t,1}.\end{align}}} The constraint $u_{t|t,m}=u_{t|t,1}$ is called the \textit{non-anticipatory} constraint. It enforces that the control applied at time $t$ is independent of scenario $m$, encoding the fact that the AV doesn't know the \textit{true modes} of the $V$ TVs at time $t$. The problem has $O(N MV)$ decision variables $\boldsymbol{\theta}_t$ but exponentially growing $O(N M^V)$ constraints, arising from the multi-modal AV predictions and collision avoidance constraints. Since the solution to \eqref{opt:MPC_skeleton} is sought for real-time applications at frequencies $\geq 10 Hz$, complex scenarios with many obstacles and modes can render the computation of \eqref{eq:MPC} impractical. 

The goal of this article is to accelerate the solution of \eqref{opt:MPC_skeleton} for real-time applications. In practice, not all collision avoidance constraints need to be enforced for solving the optimization problem \eqref{opt:MPC_skeleton}. We use Lagrange duality to interpret which TVs and their corresponding modes \textit{interact} with the AV along the prediction horizon. Leveraging this insight, we use imitation learning to train an attention-based RNN architecture for predicting which TVs, and which modes should be considered in \eqref{opt:MPC_skeleton}.



%% file: Sections/HMPC.tex
\section{Duality-based Interaction Prediction using Imitation Learning}
\label{sec:methods}

\subsection{Convex MPC formulation and Duality}\label{sec:duality-screening}

All the chance constraints can be deterministically reformulated as second-order cone (SOC) constraints in $\boldsymbol{\theta}_t$ because the process noise $w_t$ and $n^i_{k|t,j}$ are assumed to be Gaussian\cite[Chapter 5]{BEN:09}, \cite{nair2022collision}. We write the MPC optimization problem compactly as the following SOCP:
{\small{\begin{align}\label{opt:Primal_MPC}
\begin{aligned} \min_{\boldsymbol{\theta}_t}~~~&\frac{1}{2}||\mathcal{Q}_t\boldsymbol{\theta}_t||_2^2+\mathcal{C}^\top_t\boldsymbol{\theta}_t\\
\text{s.t}&~\mathcal{A}_t\boldsymbol{\theta}_t+\mathcal{R}_t\in\mathbb{K}:=(\bigotimes_{s=1}^{n_c}\mathbb{K}_s)\times \mathbb{K}_{xu}
\end{aligned}
\end{align}}}where $\mathcal{A}_t, \mathcal{R}_t, \mathcal{Q}_t, \mathcal{C}_t$ represent the predictions models and constraints, $n_c= N\cdot M^V$, $\mathcal{Q}_t\succ 0 $ and $\mathbb{K}_s$, $\forall s\in\mathbb{I}_1^{n_c}$ are the cones corresponding to the collision avoidance and $\mathbb{K}_{xu}$ is the cone that corresponds to all the state-input constraints. The exponential number of state-input constraints in \eqref{eq: xu_constr} can be reduced to $O(NMV)$ for $\mathbb{K}_{xu}$ by employing standard constraint-tightening techniques from robust optimization (Appendix \ref{app:xu_tight}). For the  $O(N M^V)$ collision avoidance constraints, we would like to solve a reduced version of \eqref{opt:Primal_MPC} involving only the \textit{necessary} collision avoidance constraints. We quantify this necessity using Lagrange duality. Consider the dual problem of \eqref{opt:Primal_MPC},
\vspace{-1mm}
{\small{\begin{align}\label{opt:Dl1_MPC}
\begin{aligned} \min_{\substack{\boldsymbol{\mu}_t, \boldsymbol{\eta}_t}}\;&[\boldsymbol{\mu}^\top_t\ \boldsymbol{\eta}^\top_t]\mathcal{R}_t+\frac{1}{2}||\mathcal{Q}_t^{-1}(\mathcal{A}_t^\top[\boldsymbol{\mu}^\top_t\ \boldsymbol{\eta}^\top_t]^\top-\mathcal{C}_t)||_2^2\\
\text{s.t}&~\boldsymbol{\mu}_t\in\bigotimes_{s=1}^{n_c}\mathbb{K}^*_s,\boldsymbol{\eta}_t\in\mathbb{K}^*_{xu}\end{aligned}
\end{align}}}

Notice that the conic feasible set of the dual problem is independent of time $t$, and is comprised of the standard SOCs and positive orthants, for which closed-form projections can be computed efficiently \cite{ali2017semismooth}.
If strong duality holds, then the optimal primal and dual solutions $(\boldsymbol{\theta}^\star_t, \boldsymbol{\mu}^\star_t, \boldsymbol{\eta}^\star_t)$ satisfy the KKT conditions:
{\small{
\begin{align}  &\mathcal{Q}^2_t\boldsymbol{\theta}^\star_t+\mathcal{C}_t-\mathcal{A}_t^\top\begin{bmatrix}\boldsymbol{\mu}^\star_t\\\boldsymbol{\eta}^\star_t\end{bmatrix}= 0\nonumber\\
&\mathcal{A}_t\boldsymbol{\theta}^\star_t+\mathcal{R}_t\in\mathbb{K},~~\boldsymbol{\mu}^\star_t\in\bigotimes_{s=1}^{n_c}\mathbb{K}^*_s,~~~\boldsymbol{\eta}^\star_t\in\mathbb{K}^*_{xu}\nonumber\\
& [\mathcal{A}_t\boldsymbol{\theta}^\star_t+\mathcal{R}_t]_{xu}\circ\boldsymbol{\eta}^\star_t = 0, \nonumber\\
&[\mathcal{A}_t\boldsymbol{\theta}^\star_t+\mathcal{R}_t]_s\circ[\boldsymbol{\mu}^\star_t]_s = 0~~\forall s\in\mathbb{I}_1^{n_c}
\end{align}}}where the $[\boldsymbol{\mu}^\star_t]_s$ denotes dual variable belonging to $s$th cone $\mathbb{K}^*_s$,
and $[\mathcal{A}_t\boldsymbol{\theta}^\star_t+\mathcal{R}_t]_{xu}$ denotes the rows of $\mathcal{A}_t\boldsymbol{\theta}^\star_t+\mathcal{R}_t$ that correspond the state-input constraints. 
Observe from the last equation that  if $[\boldsymbol{\mu}^\star_t]_s=0$, then $[\mathcal{A}_t\boldsymbol{\theta}^\star_t+\mathcal{R}_t]_s\in\text{int}(\mathbb{K}_s)$ and this constraint can be eliminated from \eqref{opt:Primal_MPC}.

Now suppose that we are given a dual feasible candidate $\hat{\boldsymbol{\mu}}_t\in\bigotimes_{s=1}^{n_c}\mathbb{K}^*_s$. Let $S^i[\hat{\boldsymbol{\mu}}_t]$ denote the duals corresponding to collision avoidance constraints with vehicle $i$,  $S_m[\hat{\boldsymbol{\mu}}_t]$ denote the duals corresponding to collision avoidance constraints in a particular mode configuration scenario $m$ and $[\mathcal{A}_t\boldsymbol{\theta}+\mathcal{R}_t]_{xu}$ be the rows corresponding to the state-input constraints. We will use ideas from sensitivity analysis of conic programs to eliminate vehicles, and scenarios to construct a reduced MPC problem as follows. 

Define the maximum possible violation of the collision avoidance constraints across all vehicles for an AV trajectory that satisfies state and input constraints,
{\small{\begin{align*}
    \bar{D}: = \max_{i, \boldsymbol{\theta}^i }\min_{\boldsymbol{\delta}^i}&~~~~ ||S^i[\mathcal{A}_t\boldsymbol{\theta}^i+\mathcal{R}_t]-\delta^i||_2\\
    \text{s.t}&~~  i \in \mathbb{I}_1^V, [\mathcal{A}_t\boldsymbol{\theta}^i+\mathcal{R}_t]_{xu}\in\mathbb{K}_{xu}, \delta^i\in S^i[\mathbb{K}]
\end{align*}}}
The quantity $\bar{D}$ is guaranteed to be finite because (i) the AV's trajectories  and target vehicles' predictions are bounded by dynamical and actuation constraints, and (ii) the vehicle geometries are compact sets \footnote{$\bar{D}$ does not need to be explicitly computed and any $\tilde{D} > \bar{D}$ is sufficient, e.g., a conservative choice could be $\tilde{D}=N\cdot M\cdot W$, where $W=$ largest dimension of drivable area. }. Also let $\delta$ be an acceptable deviation in the optimal cost of \eqref{opt:MPC_skeleton}. By using the \textit{shadow price} \cite[Chapter 5]{boyd2004convex} interpretation of the duals $S^i[\hat{\boldsymbol{\mu}}_t]$, if $||S^i[\hat{\boldsymbol{\mu}}_t]||\leq \frac{\delta}{\bar{D}}$, then the optimal cost changes  by $\approx\frac{\delta}{\bar{D}}\cdot\bar{D}=\delta $ if the collision avoidance constraints for vehicle $i$ are violated by the amount $\bar{D}$. Thus, vehicle $i$ can be ignored without significantly affecting the cost of the computed motion plan. Similarly if $||S_m[\hat{\boldsymbol{\mu}}_t]||\leq \frac{\delta}{\bar{D}\cdot V}$, the collision avoidance constraints for all the $V$ vehicles in scenario $m$ can be ignored without significantly affecting the cost, and thus, the $m$th scenario can be pruned.  

\subsection{Recurrent Attention for Interaction Duals Network (RAID-Net)}
The optimal dual variables $\boldsymbol{\mu_t^\star}\in\bigotimes_{s=1}^{n_c}\mathbb{K}^*_s$ contain information about active constraints of the primal optimization problem \eqref{opt:Primal_MPC} predicted at time $t$. The continuous vector can be converted into a binary vector $\boldsymbol{\Tilde{\mu}_t^\star} \in \{0,1\}^{n_c}$ by applying the screening rules described in \ref{sec:duality-screening}. If $[\boldsymbol{\mu_t}]_s > 0$, then the corresponding constraint is active and $[\boldsymbol{\Tilde{\mu}_t^\star}]_s=1$. Otherwise, $[\boldsymbol{\Tilde{\mu}_t^\star}]_s=0 \quad \forall s \in\mathbb{I}_1^{n_c}$.

We propose Recurrent Attention for Interaction Duals Network (RAID-Net) that predicts the dual class $\boldsymbol{\Tilde{\mu}_t}$ given the observation ($ob_t$) of the environment. The observation may contain continuous or discrete information about the environment that correlates with the dual variables. For instance, the current state of the AV, target vehicles' positions and velocities, and semantic information such as target vehicles' behavior prediction. 

\subsubsection{Expert Dataset}
With access to a simulation environment, the optimal dual solutions $\{\boldsymbol{\mu_t^\star}\}_{t=0}^T$ (i.e. expert actions in imitation learning nomenclature) are obtained by solving \eqref{opt:Primal_MPC} at each step $t$ along the trajectory with length $T$. Then, it is converted into the binary class labels $\{\boldsymbol{\Tilde{\mu}_t^\star}\}_{t=0}^T$. The optimal dual class labels and observation of the environment pairs are stored into a dataset $\mathcal{D}=\{(ob_t,\boldsymbol{\Tilde{\mu}_t^\star})\}_{t=0}^{B}$, where $B$ is the user-defined dataset size.

\subsubsection{Input Normalization and Encoding}
We assume that the observation of the environment is available and includes information about the AV's states, previous control input, and target vehicles' states along with semantic information about the autonomous and target vehicles' modes. 

The observation is normalized using environment and system information such as maximum states and inputs (restricted by the environment and system limitations), the maximum number of modes per vehicle, etc. In the context of complex interactive autonomous driving scenarios, employing graph encoding of the scene is crucial as it allows for a comprehensive representation of the intricate relationships and interactions among various entities within the scene. Nodes in the graph represent vehicles including the AV, while directed edges (originating from the autonomous vehicle) capture the dynamic relationships and interaction between the autonomous vehicle and target vehicles. We utilize an ego-centric graph encoding using time-to-collision (TTC) between the autonomous vehicle and the target vehicles to represent the edges. In RAID-Net the observation is separated per $i$-th vehicle: $ob_t^i \quad \forall i \in \mathbb{I}_0^V$, where $i=0$ represents the AV and augmented with the TTC graph encoder. Subsequently, a transformer-based encoder embeds the augmented observations, producing a scene representation feature vector denoted as $f_i \in \mathbb{R}^{d_{em}}$. In this work, we define $d_{em}= 2 \times |ob|$, where $|ob|$ represents the dimension of the observation space.

The encoder exhibits permutation invariance, indicating that the network's output remains unchanged even when the order of the input is interchanged. This invariance is achieved by summing the feature vectors $\{f_i\}_{i=0}^V$ over all vehicles before feeding them into the decoder as depicted in Fig. \ref{fig:alg_architecture}.

\subsubsection{Network Architecture}
\begin{figure*}
    \centering
    \includegraphics[width=0.7\textwidth]{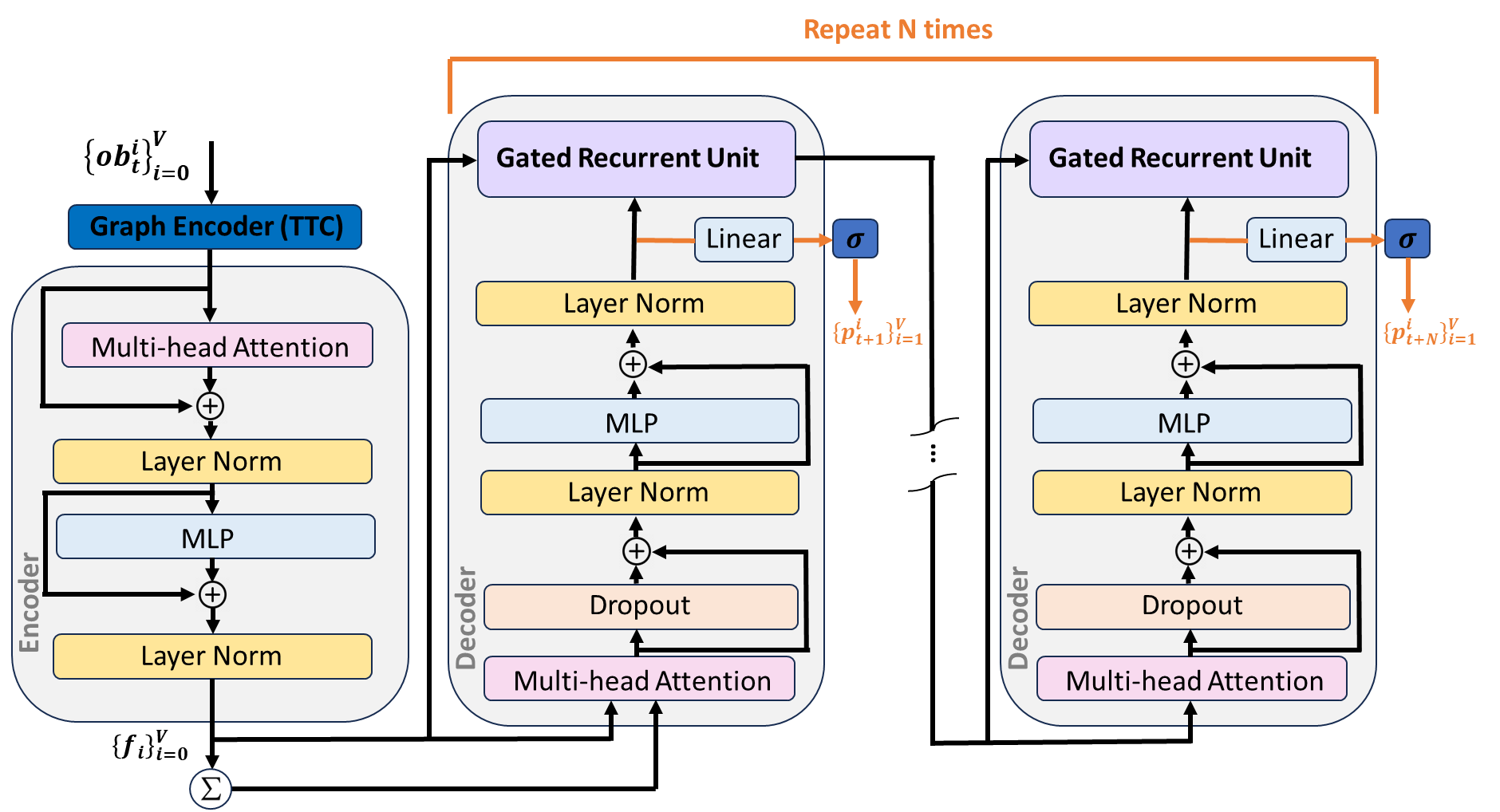}
    \caption{\small{Schematic of our Recurrent Attention for Interaction Duals Network (RAID-Net) for predicting relevant constraints for the MPC optimization problem. RAID-net is invariant to the number and order of target vehicles, and has a MPC horizon-independent memory footprint because of its recurrent architecture.}}
    \label{fig:nn_architecture}
\end{figure*}

The normalized and encoded inputs undergo $N$ sequential passes through an attention-based and gated recurrent network of decoders (the decoders share parameters), as illustrated in Fig. \ref{fig:alg_architecture}.

The decoder includes a multi-head attention mechanism to capture the different types of relationships between the encoded inputs \cite{vaswani2023attention}. Each head of the attention mechanism focuses on different interactions between AV and target vehicles. See \cite{vaswani2023attention} for detailed explanations for a multi-head attention network architecture. The multi-head attention network in the decoder features an embedding dimension of $d_{em}$ and $n_h$ denotes the number of heads. 

The decoder also features dropout and layer normalization layers to prevent overfitting and stabilize the training process \cite{dropout}\cite{layernorm2016}. A multi-layer perceptron (MLP) or a fully-connected neural network is included to allow the model to learn complex interactions and relationships in the data in the embedded dimension. The MLP in RAID-Net decoder is 
\begin{equation}
   {\small{ z_\ell = \sigma^{\text{LeakyReLU}}(W_\ell z_{\ell-1} + b_\ell), \quad \ell = 1, ... , L}}
\end{equation}
where $W_\ell, b_\ell$ the parameters of $\ell$-th layer of the network with hidden state dimension $h_{MLP}$ where the input and output of the MLP are $z_0, z_L \in \mathbb{R}^{d_{em} \cdot (V+1)}$; $\sigma^{\text{LeakyReLU}}(x)=\max(0, x) + \alpha \min(0, x)$ where $\alpha$ is the negative slope parameter that controls the activation in the negative input region; $L$ is the number of layers. The hidden states and the input to the decoder network are passed to a gated recurrent unit (GRU) which applies an input-dependent gating mechanism to modify the hidden state that will be passed on to the subsequent decoder (See Fig. \ref{fig:nn_architecture}). The GRU's input is identical to the corresponding decoder's input denoted as $z\in\mathbb{R}^{d_{em}}$. The number of features or hidden size of the GRU is set to $d_{em}\cdot(V+1)$ which represents the sequence of embedded states of all vehicles including the AV that will be passed to the next recurrent decoder. The recurrent structure is popular in obstacle motion prediction as it captures the time-dependence mechanism of motion \cite{salzmann2021trajectron, ivanovic2021mats}. Similarly, it is useful for complex interaction predictions to capture the time-dependence of the autonomous vehicle's interactions with the target vehicles and prevents the vanishing gradient problem that is prevalent in standard recurrent neural networks \cite{cho2014gru}. Also, the recurrent structure renders the RAID-Net independent of the prediction horizon ($N$) as decoders share parameters.   
A linear projection layer ($W_p z + b_p$), defined by the parameters $W_p\in \mathbb{R}^{n_c/N \times d_{em}\cdot (V + 1)}$, $b_p \in \mathbb{R}^{n_c/N} $ with sigmoid activation, lifts the sequence of embedded states to predict the class-1 probabilities of dual variables corresponding to the collision avoidance constraints with target vehicle $i$ at time $t+1$: $\boldsymbol{p^i_{t+1}} \in \mathbb{R}^{M}$, where the $m$-th element corresponds to the class-1 probability for the $m$-th mode configuration of the target vehicles in the scenario. Collecting the $\{\boldsymbol{p^i_{t+1}}\}_{i=1}^V$ from each decoder, we get 
\begin{equation}
    \boldsymbol{p_t} = [\{\boldsymbol{p^i_{t+1}}\}_{i=1}^{V}, ... , \{\boldsymbol{p^i_{t+N}}\}_{i=1}^{V}] \in \mathbb{R}^{n_c}
\end{equation}

\subsubsection{Loss function}
The binary cross-entropy loss function for training the neural network to predict the dual classes is
\begin{align}  \label{eq:loss_fn}
\ell(\boldsymbol{p},\boldsymbol{\Tilde{\mu}^\star})= 
        -\tfrac{1}{n}\sum\limits_{i=1}^n \sum\limits_{s=1}^{n_c}(w_p [\boldsymbol{\Tilde{\mu}}^\star_{i}]_s \log [\boldsymbol{p}_i]_s + \nonumber \\ (1-[\boldsymbol{\Tilde{\mu}}^\star_{i}]_s) \log(1-[\boldsymbol{p}_i]_s)], 
\end{align}
where $[\boldsymbol{p}_{i}]_s$ and $[\boldsymbol{\Tilde{\mu}^\star}_{i}]_s$ are $s$-th component of the dual class-1 probability prediction and the $s$-th component of the target class label of the $i$-th sample, respectively. $n$ is the training batch size and $w_p$ is the weight corresponding to positive class. By choosing $w_p$ appropriately, the learned classifier can conservatively over-predict class 1 and account for a potential imbalance in a dataset (i.e. number of class 0 $\gg$ number of class 1). \textit{This is critical for safety as increasing $w_p$ will reduce false negative classification of active constraints.}



\subsection{Hierarchical MPC with RAID-Net}
For online deployment, we use the RAID-Net in a hierarchical structure with the low-level MPC planner as visualized in Fig. \ref{fig:alg_architecture} and formalized in Algorithm \ref{alg:proposed_arch}. 

The RAID-Net classifier to predict the dual classes is  
\begin{equation} \label{eq:pred_rule}
\pi^{\text{RAIDN}}(\boldsymbol{\Tilde{\mu}}_t|ob_t) = 
    \begin{cases}
         [\boldsymbol{\Tilde{\mu}_t}]_s = 1 & \forall s \in\mathbb{I}_1^{n_c}\text{ if } [\boldsymbol{p_t}]_s \geq 0.5\\
        [\boldsymbol{\Tilde{\mu}_t}]_s = 0 & \text{otherwise} 
    \end{cases}
\end{equation}
where $\boldsymbol{p_t}$ is the RAID-Net output representing the probabilities of the dual classes being 1, given the observation of the environment at time $t$. Next, we define a set that contains all the indices of constraints that are predicted to be active. 
\begin{equation} \label{eq:active_set}
     \hat{\mathbb{S}}:=\{s\in \mathbb{I}_1^{n_c}|~ [\pi^{\text{RAIDN}}(\boldsymbol{\Tilde{\mu}}_t|ob_t)]_s=1\}
\end{equation} 
To recover a dual feasible candidate $\hat{\boldsymbol{\mu}}_t$, we solve the reduced linear system via least-squares,
\begin{align}{\small{
[\mathcal{A}_t]_{\hat{\mathbb{S}}\cup xu}\mathcal{Q}^{-1}_t[\mathcal{A}_t]_{\hat{\mathbb{S}}\cup xu}^\top\begin{bmatrix}[\bar{\boldsymbol{\mu}}_t]_{\hat{\mathbb{S}}}\\{\boldsymbol{\eta}}_t\end{bmatrix}\!=\![\mathcal{A}_t]_{\hat{\mathbb{S}}\cup xu}\mathcal{Q}^{-1}_t\mathcal{C}_t}}\!-\![\mathcal{R}_t]_{\hat{\mathbb{S}}\cup xu}
\end{align} to obtain the unconstrained minimizer, $\bar{\boldsymbol{\mu}}_t$, of the \textit{reduced} dual problem (by eliminating all duals not in $\hat{\mathbb{S}}$). Then we project $[\bar{\boldsymbol{\mu}}_t]_{\hat{\mathbb{S}}}$ onto the corresponding cones to get $[\hat{\boldsymbol{\mu}}_t]_{\hat{\mathbb{S}}}$ and set the other elements of $\hat{\boldsymbol{\mu}}_t$ to be $0$. Finally, we perform the sensitivity-based vehicle and scenario selection from Section~\ref{sec:duality-screening} to get
\begin{equation} \label{eq:sense_active_set} {\small{\mathbb{S}:=\hat{\mathbb{S}}\bigcap_{\substack{\forall i\in\mathbb{I}_1^V: \ ||S^i[\hat{\boldsymbol{\mu}}_t]||_2 > \frac{\delta}{\bar{D}} }}\!\mathbb{S}^i\bigcap_{\substack{\forall m\in\mathbb{I}^{M^V}_1: \ ||S_m[\hat{\boldsymbol{\mu}}_t]||_2 > \frac{\delta}{\bar{D}\cdot V}}}\mathbb{S}_m,}}
\end{equation} 
where $\mathbb{S}^i, \mathbb{S}_m\subset \mathbb{I}_1^{n_c}$ are the index sets of collision avoidance cones corresponding to vehicle $i$ and scenario $m$, respectively.
Thus, we can now define the reduced optimization problem

\begin{align}\label{opt:reduced_mpc}
&(P_{r}):\!
&{\small{\begin{aligned} \min_{\boldsymbol{\theta}_t}~~~&\frac{1}{2}||\mathcal{Q}_t\boldsymbol{\theta}_t||_2^2+\mathcal{C}^\top_t\boldsymbol{\theta}_t\\
\text{s.t}&~[\mathcal{A}_t\boldsymbol{\theta}_t+\mathcal{R}_t]_{\mathbb{S}\cup xu}\in\mathbb{K}:=(\bigotimes_{s\in\mathbb{S}}\mathbb{K}_s)\times \mathbb{K}_{xu}
\end{aligned}}}
\end{align}
The Reduced MPC (ReMPC) is given by the optimal solution of \eqref{opt:reduced_mpc} as {\small{
\begin{align}\label{eq:RMPC}u_t=\pi_{\mathrm{\text{ReMPC}}}( \mathcal{A}_t, \mathcal{R}_t,\mathcal{Q}_t,\mathcal{C}_t,\mathbb{S})=u^\star_{t|t,1}.\end{align}}}
where $\mathcal{A}_t$ and $\mathcal{R}_t$ represent the initial conditions, state and input constraints, respectively. We assume $ob_t$ is available (from the simulation environment or on-vehicle sensory devices) and contains the necessary information to construct $\mathcal{A}_t, \mathcal{R}_t, \mathcal{Q}_t, \mathcal{C}_t$. 

\begin{algorithm}
\caption{\textbf{Hierarchical MPC (HMPC):} Hierarchical Motion Planning with Duality-based Interaction Prediction}\label{alg:proposed_arch}
\begin{algorithmic}
\Require $\pi^{\text{RAIDN}}$
\State Set the task horizon $T$
\While{$t < T$}
    \State $ob_t \gets$ Observation from the environment 
    \State $\mathcal{A}_t, \mathcal{R}_t, \mathcal{Q}_t, \mathcal{C}_t \gets ob_t$ \Comment{Constructed from observation}
    \State $\hat{\mathbb{S}} \gets$ Computed using $\eqref{eq:active_set}$ from $\pi^{\text{RAIDN}}(\boldsymbol{\Tilde{\mu}_t}|ob_t)$
    \State $\mathbb{S} \gets$ Computed from sensitivity criterion \eqref{eq:sense_active_set}
    \State $u_t \gets \pi_{\mathrm{\text{ReMPC}}}( \mathcal{A}_t, \mathcal{R}_t,\mathcal{Q}_t,\mathcal{C}_t,\mathbb{S})$
    \State Apply $u_t$ to system
    \State $t \gets t+1$
\EndWhile
\end{algorithmic}
\end{algorithm}

%% file: Sections/traffic_intersection.tex
\section{Example: Planning at a Traffic Intersection}
\label{sec:example}

\subsection{Simulation Environment}
\begin{figure}
    \centering
    \includegraphics[width=0.88\columnwidth]{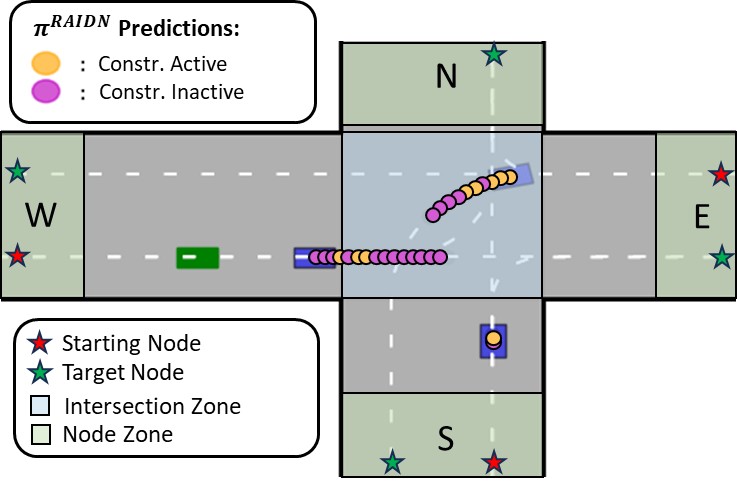}
    \caption{\small{An example scene in the custom unsignalized intersection environment. The autonomous vehicle (green rectangle) is approaching the intersection and interacts with the target vehicles (blue rectangles).The active and inactive constraint predictions from the $\pi^{\text{RAIDN}}$ are depicted as yellow and magenta ellipses, respectively.}}
    \label{fig:sim_env}
\end{figure}

We created a customized Python simulation environment for unsignalized traffic intersection based on OpenAI gymnasium \cite{openai_gymnasium}. The simulation environment has four node zones: \textit{N, S, W, and E} as shown in Fig. \ref{fig:sim_env}. From each node zone, vehicles are randomly generated in the following order: Firstly, the autonomous vehicle is placed at the starting node in \textit{W} zone, starting with a random target node (\textit{N} or \textit{E} defining 2 modes for autonomous vehicle). Note that the \textit{S} node isn't considered a target for the vehicle starting in \textit{W} zone as it leads to minimal interaction between vehicles.
Next, $V$ target vehicles are randomly spawned. $V$ is a random integer from the interval [1,$V_{max}$], where $V_{max}$ is the maximum possible number of target vehicles defined by the user. In this work, we arbitrarily set $V_{max}=3$. 

Each of these target vehicles originates from distinct zones (\textit{W, S, and E}) and is assigned a randomly determined target node based on the available modes outlined in Table~\ref{tab:mode_table}. The target vehicle spawning from the \textit{E} zone, serving as cross-traffic for the autonomous vehicle, is provided with four modes for diverse interactive scenarios, while target vehicles starting from other zones have two modes each. The total possible mode configuration of the target vehicles in the scenario is then $M=16$, which leads to $13\cdot16\cdot3=624$ collision avoidance constraints for \eqref{opt:Primal_MPC} when $N=14$.

Given the starting and target node, each vehicle follows a fixed reference trajectory marked with white dashed lines in Fig. \ref{fig:sim_env}. The vehicle states are $x=[s,v]\in\mathbb{R}^2$, where $s$ and $v$ are displacement along the reference trajectory and velocity, respectively. The target vehicle starting from the \textit{W} node begins 8 meters ahead of the autonomous vehicle, moving at an initial velocity of 8 $m/s$. Target vehicles from the \textit{E} zone start with an initial velocity of 8 $m/s$  (7 if slow), while those from the \textit{S} zone begin with 7 $m/s$.

\begin{table}[!ht]
    \centering
    \caption{\small{Simulator Vehicle's Modes}}

    \label{tab:mode_table}    
    \begin{tabular}[c]{| c | c | c|}
        \hline
            Start Node & Target Node & $\#$ Modes \\
        \hline
        \hline
       \textit{W}  & \textit{E}, \textit{N} &  2 \\
        \hline
       \textit{S}  & \textit{N, E} &  2 \\
        \hline
       \textit{E}  & \textit{W, W }(slow), \textit{S, N} &  4 \\
        \hline
    \end{tabular}
\end{table}

The target vehicle resets to its starting node upon reaching its target node, ending the simulation episode if a collision occurs or when the autonomous vehicle reaches its target node. Target vehicles employ a Frenet-frame Intelligent Driver Model (fIDM) for safe and human-like acceleration, considering virtual projections of surrounding vehicles on the controlled vehicle's reference trajectory, akin to the generalized Intelligent Driver Model in \cite{gIDM}. The fIDM also features logic gates that prioritize vehicles within intersections, emulating human-like driving behavior. For instance, the approaching vehicle will yield to the surrounding vehicle that is already in the interaction zone as shown in Fig. \ref{fig:sim_env}. After computing acceleration inputs for the autonomous vehicle and target vehicles using the relevant motion planner and fIDM, they are simulated forward by a time step ($\Delta t$) along their respective reference trajectories using the kinematic bicycle model.

The observation vector available to the ego vehicle from the simulator is represented as $ob_{t} = [x_t, u_{t-1}, mode_{ev}, \{o^i_t\}^{V_{max}}_{i=1}, \{mode_{i}\}_{i=1}^{V_{max}}, \{TTC_i\}_{i=0}^{V_{max}} ]\in \mathbb{R}^{17}$, where $x_t$, $u_{t-1}$, and $mode_{ev}$ denote the ego states, control input at previous time step, and EV's mode, respectively. $o^i_t$ represents the state of the $i$-th target vehicle, $mode_i$ denotes its mode, and $TTC_i$ denotes the time-to-collision between the autonomous vehicle and the $i$-th target vehicle, with $TTC_0=0$. When $V<V_{max}$, we spawn dummy vehicles outside the traffic scene for the rest.

\subsection{RAID-Net Imitation Learning}
We use a behavior cloning algorithm \cite{Michie1990CognitiveMF-bc} to train the RAID-Net to mimic the expert (optimal dual solutions to \ref{opt:Primal_MPC}). Data is collected by rolling out trajectories in the aforementioned simulator using the expert policy \eqref{opt:Primal_MPC} with prediction horizon $N=14$ and sampling time $\Delta t=0.2 s$. Then, $(ob_t,\boldsymbol{\Tilde{\mu_t}^\star})$ pairs at each time step $t$ are recorded. During the data collection phase, the optimization problem is formulated in CasADi\cite{casadi} and solved using IPOPT\cite{ipopt} to extract optimal dual solutions. Repeating the process initialized with randomly generated scenarios, 120,315 data points are collected. We decompose the dataset into training (85$\%$) and test (15$\%$) datasets. 

In dual-class classification for predicting active constraints, minimizing false negatives is crucial for safety. While low precision is acceptable, high recall is desired, which represents the model's ability to find all positive classes. To counter dataset imbalance and introduce a positive-class bias in the model, we employ $w_p = 4$ in \eqref{eq:loss_fn} during model training. Additionally, we address dataset imbalance using a weighted random sampler in PyTorch during training, assigning a higher weight to the under-represented class to achieve balanced class representation in the training batch.
\begin{table}[!ht]
    \centering
    \caption{\small{RAID-Net model parameters}}

    \label{tab:params}    
    \begin{tabular}[c]{|c | c | c | c| c|}
        \hline
            $\alpha$ & $p_{\text{dropout}}$& $h_{\text{MLP}}$ & L & $n_h$ \\
        \hline
        \hline
       -0.1  & 0.1 & 128& 6& 1\\
        \hline
    \end{tabular}
\end{table}
The RAID-Net is constructed using PyTorch and trained using the training dataset and the loss function \eqref{eq:loss_fn}. The parameters used to construct the RAID-Net are reported in Table~\ref{tab:params}.
We train the RAID-Net for 3000 epochs using a batch size of 1024 and the Adam with a constant learning rate of 0.001 and $\beta_1, \beta_2 = (0.9,0.99)$ \cite{adamw}. The training process takes in total of 4 hours. \footnote[2]{Training and experiments were run on a computer with a Intel i9-9900K CPU, 32 GB RAM, and a RTX 2080 Ti GPU. \label{footnote_1}}

%% file: Sections/results.tex
\section{Results}
\label{sec:results}

 

In this section, we first present the imitation learning results of the RAID-Net. Secondly, we compare the performance of our proposed \textbf{HMPC} policy \ref{alg:proposed_arch} to the full MPC policy \eqref{opt:Primal_MPC} ($N=14, \Delta t=0.2 \;\text{s}$ for both policies) across 100 randomly generated scenarios in numerical simulations \footref{footnote_1}. During the evaluation of HMPC and full MPC policies, the optimization problems are formulated in CasADi and the MPC SOCPs are solved using Gurobi \cite{gurobi}. 

\subsection{RAID-Net Evaluation}\label{sec:classifier_eval}
\begin{figure}
    \centering
    \includegraphics[width=0.9\columnwidth]{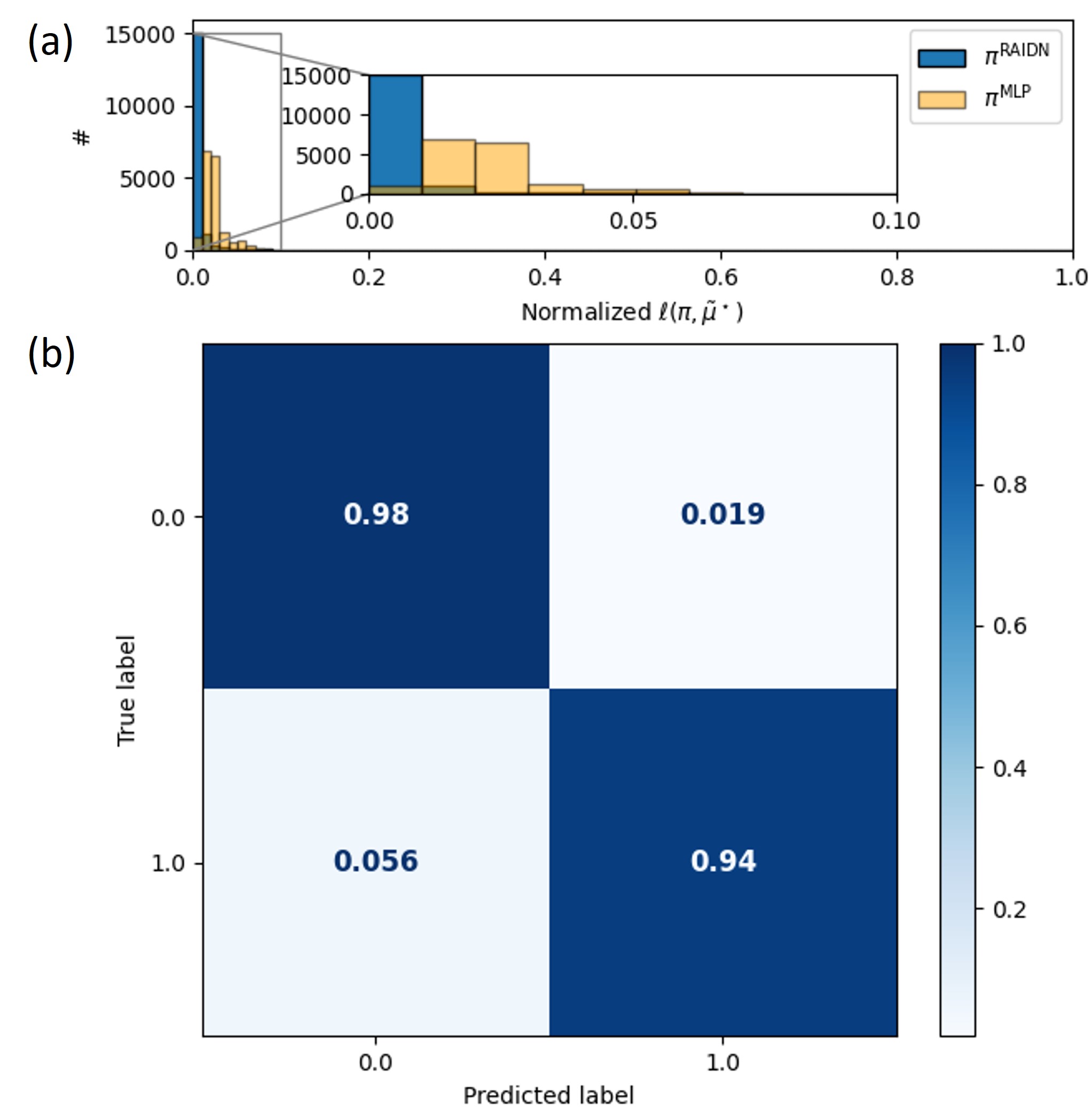}
    \caption{\small{RAID-Net evaluation results on a test dataset: (a) histogram of the normalized loss value of RAID-Net versus MLP neural network, (b) normalized confusion matrix of RAID-Net}}
    \label{fig:model_evaluation}
    \vspace{-0.5cm}
\end{figure}

The trained RAID-Net model was evaluated on the test dataset. We compare the normalized loss statistics of RAID-Net to an MLP NN ($\pi^{\text{MLP}}$) trained with the configuration as RAID-Net and following parameters: $h_{\text{MLP}}=128, L=6$ in Fig. \ref{fig:model_evaluation}a. We report the normalized confusion matrix of RAID-Net in Fig. \ref{fig:model_evaluation}b.
The RAID-Net model adeptly captures intricate interactions between the AV and TVs, surpassing the capabilities of an MLP NN.
Further, the RAID-Net model achieved a recall of 0.94 and a precision of 0.44—in other words, it correctly predicts an interactive constraint 44$\%$ of the time. RAID-Net training is configured to trade-off precision for high recall as false positive classifications are not critical to safety. Evaluated on the test dataset, it correctly predicts 98.1$\%$ of the total interactive duals (Class 1) 
with a low 0.056 false negative rate. This performance showcases the proposed architecture's effectiveness in capturing complex interactions between the AV and TVs with multi-modal predictions.

\subsection{Policy Comparison}\label{sec:policy_eval}
The performance metrics over 100 runs are recorded in Table~\ref{tab:comparison_tab}. We record $\%$ of time steps where MPC infeasibility and collision with a target vehicle were detected, along with average $\%$ of constraints that were imposed in the MPC. Furthermore, average times for solving \eqref{opt:Primal_MPC} and \eqref{opt:reduced_mpc} as well as the average times for querying the RAID-Net. Finally, the average task completion (reaching the target node) time for \textbf{HMPC} normalized by that of the full MPC is reported. Fig. \ref{fig:sim_env} presents a snapshot of an example scene in the custom simulation environment, visualizing the $\pi^{\text{RAIDN}}$ predictions. The \textbf{HMPC} algorithm imposes only the collision avoidance constraints corresponding to the yellow multi-modal prediction of the target vehicle. A video demonstrating the \textbf{HMPC} algorithm in multiple scenarios can be found here: {\color{blue} \href{https://youtu.be/-pRiOnPb9_c}{link}}. The code for experiments can be found here: {\color{blue} \href{https://github.com/MPC-Berkeley/hmpc_raidnet}{link}}

\begin{table}[!ht]
    \centering
    \scriptsize
    \caption{\small{Performance metrics across all policies for \ref{sec:policy_eval}.}}

    \label{tab:comparison_tab}    
    \begin{tabular}[c]{|c|c|c|}
    
        \hline
            Performance metric &
        Full MPC \eqref{opt:Primal_MPC} & HMPC \eqref{alg:proposed_arch} \\
        \hline
        \hline
       Feasibility (\%)    & 98.97 & $\mathbf{99.79}$\\
       Collision (\%) & $\mathbf{0}$ & 4.0 \\
       Avg. Constraints Enforced (\%)    & 100 & $\mathbf{17.45}$ \\
         Avg. Solve Time (s)    & $0.92\pm0.18$ & $\mathbf{0.063\pm0.073}$ \\
         Avg. RAID-Net Query Time (s)    & $\mathbf{-}$ & $\mathbf{0.013\pm0.003}$ \\
         Avg. Total Computation Time (s) & $0.92\pm0.18$ & $\mathbf{0.076\pm0.076}$ \\
        Avg. normalized task completion time & \multirow{2}{*}{1} & \multirow{2}{*}{$\mathbf{0.91}$} \\ 
         w.r.t. Full MPC & &   \\
        \hline
    \end{tabular}
\end{table}

\subsubsection{Discussion} 
On average, the solve time including the RAID-Net query time for the \textbf{HMPC} algorithm is \textbf{12 times faster} than that of the full MPC as highlighted in the $6$-th row of Table \ref{tab:comparison_tab}. This significant acceleration in solving time is attributed to the reduced number of constraints using our proposed algorithm. The \textbf{HMPC} algorithm enables real-time applications at frequencies $\geq 10 Hz$ by predicting interactions and selectively imposing only safety-critical multi-modal collision avoidance constraints with multiple target vehicles, which is not feasible in real-time otherwise. Additionally, \textbf{HMPC} algorithm had slightly higher average feasibility $\%$ which is also attributed to imposing fewer constraints. Considering that only 1.52$\%$ of the total constraints were active in the test dataset, a prediction of 17.45$\%$ for active constraints is a conservative estimation that was deliberately achieved for safety. The \textbf{HMPC} algorithm achieved a 0.91 normalized task completion time, indicating a 9$\%$ faster task completion compared to control with the full MPC. The \textbf{HMPC}-AV completes the task faster as it is optimistic, imposing fewer constraints, while the full MPC-AV exhibits a more conservative behavior.

We note that out of 100 runs, the autonomous vehicle collided with a target vehicle 4 times when controlled using the \textbf{HMPC} algorithm compared to 0 times using the full MPC. Collisions occurred due to false negative classification of dual variables leading the \textbf{HMPC} algorithm to screen out incorrect constraints.

The results highlight the advantages of screening inactive constraints from the MPC formulation for motion planning in highly interactive traffic scenarios to accelerate the computation of MPC. Numerical simulations revealed that, in practice, very few constraints are active in MPC formulation for motion planning with multi-modal collision avoidance constraints. With the proposed RAID-Net architecture, the model can learn to predict which target vehicles, and which modes along the prediction horizon interact with the autonomous vehicle given the graph encoding of the scene. Thus by imposing only the relevant, interactive constraints, we achieve over 12 times improvement in solution time while ensuring safety with a high probability. 


%% file: Sections/conclusion.tex
\section{Conclusion and Future Work}
\label{sec:conclusion}
We proposed a hierarchical architecture for scalable real-time MPC for complex, multi-modal traffic scenarios, consisting of 1) RAID-Net, an attention-based recurrent NN architecture that predicts feasible dual solutions of the MPC problem for identifying relevant AV-TV interactions along the MPC horizon using a sensitivity-based criterion, and 2) a reduced Stochastic MPC problem that eliminates irrelevant collision avoidance constraints for computational efficiency. We demonstrate our approach at a simulated traffic intersection with interactive vehicles and achieve a 12x speed-up in MPC solve times. Current drawbacks of our approach are the lack of guarantees on the screened collision avoidance constraints and over-parameterized policies in the MPC. Also, RAID-Net lacks generalizability to other intersection topologies. For future work, we aim to remedy these concerns by further exploiting the duality and convexity properties of the MPC optimization problem, using DAgger \cite{dagger} to address distributional shifts in behavior cloning, and extensively testing the generalization of RAID-Net on a real-traffic dataset.
 Furthermore, we would like to apply the ideas of duality-based interaction predictions towards coordinated multi-vehicle path planning \cite{KIM2023}.

%% file: Sections/appendix.tex
\appendix
\subsection{Constraint Tightening for Reducing Multi-modal State-Input constraints}\label{app:xu_tight}
First, we enforce $O(N M V)$  joint chance constraints {\small{\begin{align}
&\mathbb{P}\left[\dfrac{-1}{V}\gamma\leq K^i_{k|t,j}(o^i_{k|t,j}-\mathbb{E}[o^i_{k|t,j}]) \leq \dfrac{1}{V}\gamma\right] \geq 1-\beta
\end{align}}}
for some $\beta < \epsilon$ and reactive control authority $\gamma\in\mathbb{R}^{n_u}$ within input constraints. Define the \textit{tightened} state-input constraints $\Tilde{\mathbb{XU}}_k =\{(x,u)| F^x_{k}x + F^u_{k}u\leq \tilde{f}_k\} $ where {\small{\begin{align*}
\tilde{f}_k &= f_k -\max_{\substack{-\gamma\leq z_l\leq\gamma\\ l =t,..,k}}F^x_{k}\sum_{l=t}^k \prod_{v=l}^kA_v B_l z_l + F^u_{k}z_k
\end{align*}}}which can be computed in closed-form using the dual norm\cite[Chapter 1]{BEN:09}. Let the nominal predicted state for time $p$ be $\bar{x}_{p|t}=\prod_{v=t}^kA_vx_t+\sum_{l=t}^p \prod_{v=l}^pA_v B_lh_{l|t}$. The quantity $\bar{f}_k$ ensures that 
 if $(\bar{x}_{k|t}, h_{k|t})\in\Tilde{\mathbb{XU}}_{K}$ and $\dfrac{-1}{V}\gamma\leq K^i_{k|t,j}(o^i_{k|t,j}-\mathbb{E}[o^i_{k|t,j}]) \leq \dfrac{1}{V}\gamma$, then $(x_{k|t,m}, u_{k|t,m})\in\mathbb{XU}_k$ for \textit{any} $m\in\mathbb{I}_1^{M^V}$. Then we replace the exponentially many probabilistic state-input constraints in \eqref{opt:MPC_skeleton} using: 
{\small{\begin{align}
    &\mathbb{P}\left[\dfrac{-1}{V}\gamma\leq K^i_{k|t,j}(o^i_{k|t,j}-\mathbb{E}[o^i_{k|t,j}]) \leq \dfrac{1}{V}\gamma\right] \geq 1-\beta,\nonumber\\ 
    &\mathbb{P}[(\bar{x}_{k+1|t},h_{k|t})\notin\tilde{\mathbb{XU}}_k]\leq\epsilon-\beta,~\forall k\in\mathbb{I}_t^{t+N}, i\in\mathbb{I}_1^V, j\in\mathbb{I}_1^M\label{eq:LHS_xu}\\
    \Longrightarrow &\mathbb{P}((x_{k+1|t,m},u_{k|t,m})\!\not\in\mathbb{XU}_k)\!\leq\! \epsilon, \forall m\in\mathbb{I}_1^{M^V}, k\in\mathbb{I}_t^{t+N}\label{eq:RHS_xu}.
\end{align}}}
Thus, the  $O(N M V)$ state-input constraints \eqref{eq:LHS_xu} can be enforced to inner-approximate the $O(N M^V)$ state-input constraints in \eqref{eq:RHS_xu}.